\documentclass[letterpaper, 10 pt, conference]{ieeeconf}

\IEEEoverridecommandlockouts        % needed to use \thanks
\usepackage{cite}
\usepackage{amsmath,amssymb,amsfonts}
\usepackage{graphicx}
\usepackage{textcomp}

\title{\LARGE \bf
Fine Wrist Control as a Marker of Surgical Teleoperation Expertise
}

\author{Mary Kate Gale$^{1}$, Shujiro Shobayashi$^{1,2}$, Sangeet Satpathy$^{1}$,
Nitsan Davidor$^{3}$, Ilana Nisky$^{3}$, Allison Okamura$^{1}$%
\thanks{This work was supported in part by the National Science Foundation Graduate Research Fellowship, the Link Foundation Modeling, Simulation, and Training Fellowship, the National Science Foundation under Grant 1828993, and the U.S.-Israel Binational Science Foundation under Grant 2023022.}%
\thanks{$^{1}$Department of Mechanical Engineering, Stanford University, Stanford, CA, USA. {\tt\small mgale7@stanford.edu}}%
\thanks{$^{2}$Robotics Program, \'Ecole Polytechnique F\'ed\'erale de Lausanne, Lausanne, Vaud, Switzerland.}%
\thanks{$^{3}$Department of Biomedical Engineering, Ben-Gurion University of the Negev, Beer Sheva, Israel.}%
}

\begin{document}

\maketitle
\thispagestyle{empty}
\pagestyle{empty}

%%%%%%%%%%%%%%%%%%%%%%%%%%%%%%%%%%%%%%%%%%%%%%%%%%%%%%%%%%%%%%%%%%%%%%%%%%%%%%%%
\begin{abstract}
Unlike most intensely physical pursuits, surgical robotic teleoperation training focuses primarily on task outcomes rather than surgeon body posture or biomechanics during task completion. Toward the question of the role of biomechanics in surgical expertise, we sought to characterize the articular motion of expert teleoperators as compared to novice users. Twenty-seven novices and nine experts completed a non-medical cylinder-on-peg transfer task while their upper limb biomechanics were recorded via motion trackers. During more difficult motions, experts stabilized their wrist motion more than novices, while maintaining adequate range of motion in their shoulder and elbow and completing the task significantly faster than novices. This marker of expertise suggests the importance of attention to user biomechanics during teleoperation of surgical robots.
\end{abstract}

%%%%%%%%%%%%%%%%%%%%%%%%%%%%%%%%%%%%%%%%%%%%%%%%%%%%%%%%%%%%%%%%%%%%%%%%%%%%%%%%
\section{INTRODUCTION}

Robotic minimally invasive surgery has become the gold standard for many surgical procedures, and clinicians must undergo years of intensive training before they are qualified to operate on patients. This training typically involves rote repetition of tasks with offline feedback and a focus on downstream outcomes, rather than user behavior that influences those outcomes \cite{b1}. 

Previous work has demonstrated differences in motor control strategies between expert and novice users of robotic surgical systems, even during simple, non-clinical movements \cite{b2}. Experts coordinate joint angles to stabilize hand movements more effectively than novices, especially when teleoperating (as opposed to freehand motion) \cite{b3}. However, previous motor control work has focused primarily on hand and end-effector movements, without much attention to the intermediate links in the chain between user intention and overall motion. 

In this work, we compared the biomechanics of expert and novice users while completing a straightforward, non-medical task and assessed the difference in joint recruitment based on experience level. We found that experts tend to move their shoulder and elbow similarly to novices but demonstrate much better wrist control during more difficult movements. 

\section{MATERIALS AND METHODS}

\subsection{Experimental Setup}

Participants completed a cylinder-on-peg transfer task. The task board consisted of four blue and white cylinders that rested on four center pegs, with eight pegs on the outside that illuminated either blue or white with embedded LEDs to indicate target cylinder placement (Figure 1A). The task board was electrified with conductive paint, allowing for real-time detection of cylinder placement and removal. During the experiment, participants wore four 6-degree-of-freedom (DOF) IMU- and vision-based motion trackers (HTC, Taoyuan, Taiwan) on the back and right arm to track articular motion (Figure 1B). The experiment was completed with a da Vinci Research Kit (dVRK) \cite{dvrk} consisting of a da Vinci S surgeon-side console and da Vinci Si patient-side arms, instrumented with bipolar fenestrated forceps.

\begin{figure}[t]
\centerline{\includegraphics[width=\columnwidth]{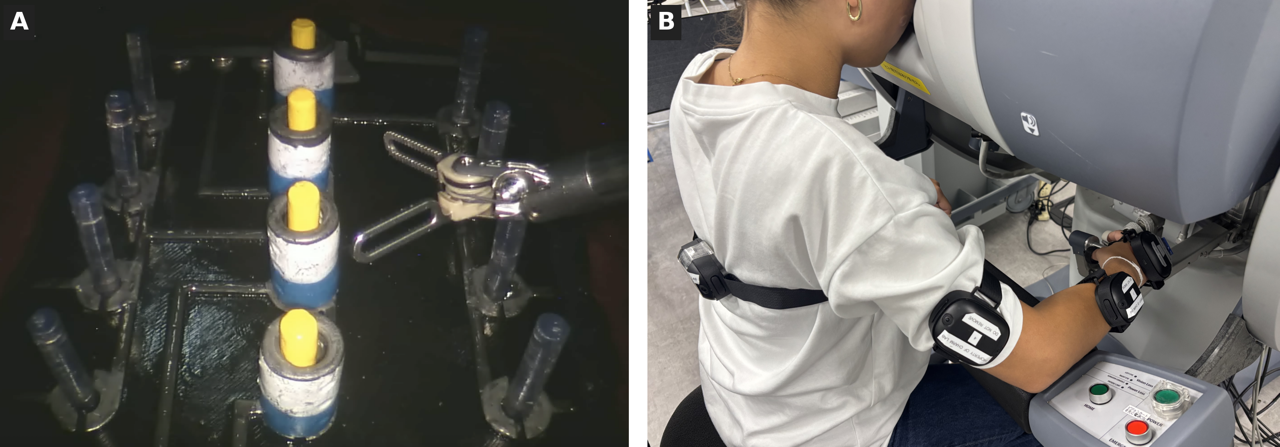}}
\caption{(A) Task board as seen through the dVRK endoscopic camera, displaying blue and white center cylinders, light-up outer pegs, and surgical instrument for grasping. (B) Location of motion trackers on participant's torso, right upper arm, right forearm, and right hand.}
\label{setup}
\end{figure}

\subsection{Methods}

The experimental protocol was approved by Stanford University’s Institutional Review Board (protocol \#22514), and all participants gave informed consent. Data were analyzed for 27 novices (no surgical or medical experience) and 9 experts (robotic surgeons with $>$250 hours of laparoscopic teleoperated surgical experience). 

The experiment consisted of forty trials, broken into two twenty-trial blocks with a three-minute break in between. At the beginning of each trial, two lights of a single color would appear on either side of a cylinder. Cylinders were always placed with the blue side down on the task board, and participants were instructed that the color of the light indicated the desired orientation of cylinder placement. When the indicated cylinder was lifted, only one light would remain on (``target" location). The participant then placed the cylinder at the target location. For trials with blue lights, the cylinder was moved directly without any rotation; for trials with white lights, the cylinder was inverted. Novice participants were instructed that it was easier to invert the forceps first before grasping for trials that necessitated inversion; experts were also given this advice, but told to use whatever method felt more natural to them. All participants were told to complete the task as quickly and smoothly as possible. If the participant dropped the cylinder or there were mechanical issues within a trial, the trial was manually reset by the experimenter and the trial timing was edited to be limited to successful movement.

Participants were only given the right surgeon-side arm of the dVRK. The workspace was tuned so that all necessary locations could be seen and reached without moving the endoscopic camera or clutching the surgeon-side arm. 

\subsection{Data Analysis}

Data streams included kinematics of the surgeon-side and patient-side robotic arms, video footage of the task, 3D location and orientation of all four motion trackers, and timing of the start and stop of each trial. The start of a trial was considered to be the point at which the cylinder was lifted, and the end was when the cylinder was placed. 

Using OpenSim's IMU Inverse Kinematics solver \cite{b4, b5}, the motion tracker movements were transformed into upper-limb joint movements (shoulder, elbow, and wrist). Joint movements were decomposed into seven degrees of freedom: wrist flexion/extension, wrist radial/ulnar deviation, forearm pronation/supination, elbow flexion/extension, shoulder plane of elevation, shoulder elevation, and shoulder axial rotation \cite{b6}. Joint movements were analyzed by trial type (blue light/non-inversion trials vs. white light/inversion trials). Within each trial, we computed the range of motion (ROM) for each joint, using the 5th percentile and 95th percentile of joint angle to define the range. A Mann-Whitney U test was used to determine the difference in average ROM between experts and novices within trial types. Significance was determined at $p<0.05$.

\begin{figure}[t]
\centerline{\includegraphics[width=\columnwidth]{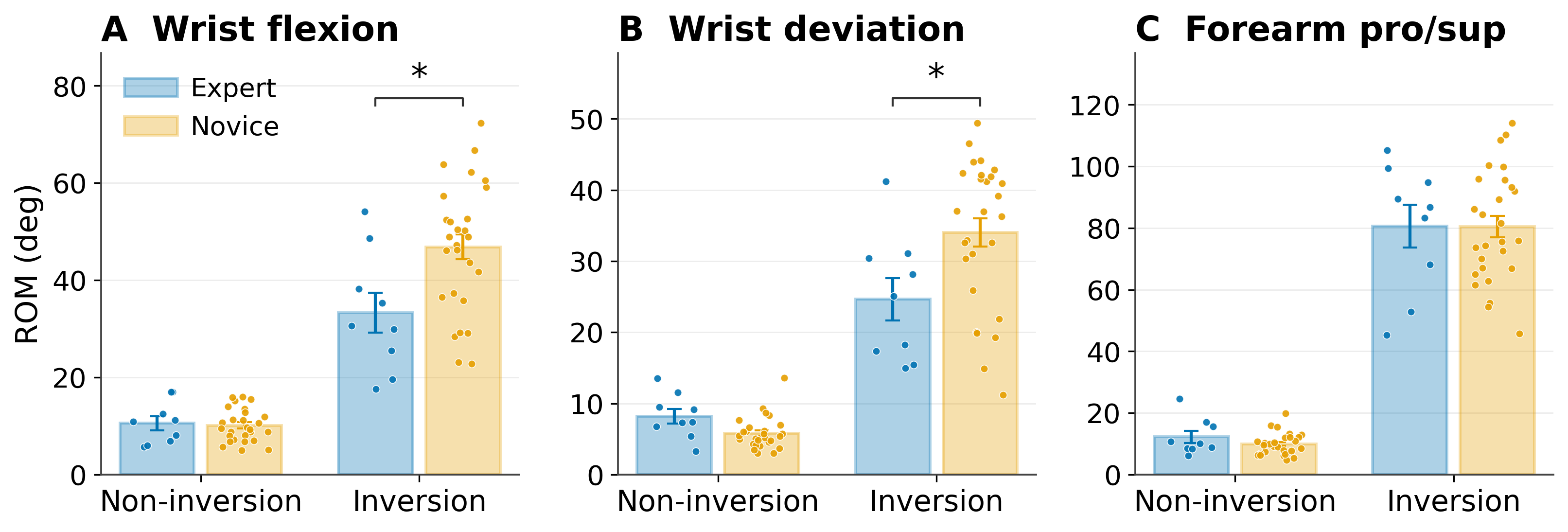}}
\caption{Expert (blue) vs. novice (orange) articular range of motion during non-inversion vs. inversion trials for (A) wrist flexion and extension, (B) wrist radial/ulnar deviation, and (C) forearm pronation and supination. Each dot represents the average range of motion for one participant.}
\label{results}
\end{figure}

\section{PRELIMINARY RESULTS AND DISCUSSION}

Experts primarily displayed behavioral differences from novices in the more difficult inversion, or white-light, trials. During these trials, novices displayed significantly higher ROMs in wrist flexion/extension (novices: $46.8^\circ$ vs. experts: $33.3^\circ$; $p=0.020$) and wrist radial/ulnar deviation (novices: $34.0^\circ$ vs. experts: $24.7^\circ$; $p=0.017$), as seen in Figure 2A-B. These differences did not appear in non-inversion trials with a simple, straight-across movement for wrist flexion/extension (novices: $10.1^\circ$ vs. experts: $10.6^\circ$; $p=0.812$). Upstream of the wrist, experts and novices displayed remarkably similar ranges of motion: in forearm pronation/supination, both experts and novices had small ranges of motion for non-inversion trials (novices: $9.9^\circ$ vs. experts: $12.2^\circ$; $p=0.391$) and large ranges of motion for inversion trials (novices: $80.4^\circ$ vs. experts: $80.6^\circ$; $p=0.898$). Similarly, both groups had markedly higher ROMs in elbow flexion/extension and shoulder elevation for inversion trials than non-inversion trials, and somewhat higher ROMs in shoulder axial rotation and plane of elevation for inversion trials. These results suggest that experts are better able to control the fine details of wrist motion during more complex motions.

Experts also completed inversion trials more quickly than novices (novices: 9.0s vs. experts: 6.7s; $p=0.012$). They tended towards more rapid trial completion in non-inversion trials, but this did not reach the level of significance (novices: 5.9s; experts: 4.8s; $p=0.093$).  

This study was limited in its assessment of naturalistic expert motion by the restrictions placed on the task to ensure it was sufficiently simple to allow novices to succeed. During the study, experts expressed frustration at having to complete the task unimanually, not being allowed to clutch, and not being able to move the camera to visualize the back of the task board. A future longitudinal study may involve more teleoperation training for novices and higher task complexity.

These preliminary results suggest strong differences in motor control strategies during teleoperation based on experience level, with experts demonstrating better control of small wrist movements. Future work could consider these differences and their implications for training of novice surgeons, through the implementation of biomechanically-aware training paradigms that focus on wrist stabilization during complex or inverted movements. The analysis of this work is ongoing, and its results may have vital implications for future training strategies for novice surgeons. 

%\addtolength{\textheight}{-2cm}   % uncomment on the page BEFORE the last page
                                   % to balance final-page column lengths

%%%%%%%%%%%%%%%%%%%%%%%%%%%%%%%%%%%%%%%%%%%%%%%%%%%%%%%%%%%%%%%%%%%%%%%%%%%%%%%%

\end{document}